\documentclass{article}

\PassOptionsToPackage{numbers, sort&compress}{natbib}

\usepackage[preprint]{neurips_2024}

\usepackage[utf8]{inputenc} 
\usepackage[T1]{fontenc}    
\usepackage{hyperref}       
\usepackage{url}            
\usepackage{booktabs}       
\usepackage{amsfonts}       
\usepackage{nicefrac}       
\usepackage{microtype}      
\usepackage{xcolor}         

\usepackage{graphicx}
\usepackage{amsmath} 
\usepackage{multirow}

\title{Self-Supervised Learning for Building Robust Pediatric Chest X-ray Classification Models}

\author{%
	Sheng Cheng \\
	Department of Computer Science\\
	Rice University\\
	Houston, TX 77005 \\
	\texttt{sc159@rice.edu} \\
	\And
	Zbigniew A. Starosolski  \\
	Department of Radiology\\
	Baylor Collage of Medicine \\
	Houston, TX 77030  \\
	\texttt{devika@rice.edu} \\
	\And
	Devika Subramanian  \\
	Department of Computer Science\\
	Rice University \\
	Houston, TX 77005 \\
	\texttt{devika@rice.edu}
}

\begin{document}

	\maketitle

	\begin{abstract}
		Recent advancements in deep learning for Medical Artificial Intelligence have demonstrated that models can match the diagnostic performance of clinical experts in adult chest X-ray (CXR) interpretation. However, their application in the pediatric context remains limited due to the scarcity of large annotated pediatric image datasets. Additionally, significant challenges arise from the substantial variability in pediatric CXR images across different hospitals and the diverse age range of patients from 0 to 18 years. To address these challenges, we propose SCC, a novel approach that combines transfer learning with self-supervised contrastive learning, augmented by an unsupervised contrast enhancement technique. Transfer learning from a well-trained adult CXR model mitigates issues related to the scarcity of pediatric training data. Contrastive learning with contrast enhancement focuses on the lungs, reducing the impact of image variations and producing high-quality embeddings across diverse pediatric CXR images. We train SCC on one pediatric CXR dataset and evaluate its performance on two other pediatric datasets from different sources. Our results show that SCC's out-of-distribution (zero-shot) performance exceeds regular transfer learning in terms of AUC by 13.6\% and 34.6\% on the two test datasets. Moreover, with few-shot learning using 10 times fewer labeled images, SCC matches the performance of regular transfer learning trained on the entire labeled dataset. To test the generality of the framework, we verify its performance on three benchmark breast cancer datasets. Starting from a model trained on natural images and fine-tuned on one breast dataset, SCC outperforms the fully supervised learning baseline on the other two datasets in terms of AUC by 3.6\% and 5.5\% in zero-shot learning.
	\end{abstract}
	
	\section{Introduction}
	
	Deep learning has significantly revolutionized pneumonia diagnosis based on Chest X-ray (CXR) images, demonstrating the potential to match the performance of clinical experts in pneumonia classification tasks. With the ability to process extensive amounts of medical imaging data, deep learning models excel in recognizing intricate patterns and abnormalities in CXRs associated with pneumonia \cite{xrv, deepCovidXR, Cross_Lingual, multimodal, seg_class}.  However, pediatric CXR images present unique challenges compared to adult CXRs due to severe image noise, varying postures, and other complexities, which can affect the performance of these models on pediatric cases.

	1. \textbf{Obstacles in transferring adult CXR models to pediatric CXR images: limited dataset and distribution shift}
	
	Developing a pediatric CXR model presents three main challenges: the scarcity of pediatric CXR images due to X-ray exposure and patient privacy concerns,  the  domain gap between adult and pediatric images, and distribution shift caused by image variations within pediatric datasets.
	Given the limited availability of pediatric datasets, leveraging models trained on extensive adult CXR datasets is appealing. However, previous studies \cite{pretraining, domainGap, domainGap2, distribution_shift} have highlighted significant domain shifts between adult and pediatric datasets, indicating that a model trained on one population does not maintain the same diagnostic performance in another, particularly between adults and children. Pediatric CXRs \cite{children2, children3} are inherently more complex than adult images due to factors such as insufficient inflation, improper positioning, non-standard exposure, clothing, and the presence of external or implanted medical devices, contributing to the adult-pediatric (AP) domain gap. Additionally, the pediatric-pediatric (PP) domain gap poses another challenge: Seyyed-Kalantari et al. \cite{sexAndAge} found that CXR datasets from multiple sources exhibit different biases, and Cruz et al. \cite{bias} suggested that these biases can stem from various factors, including image processing artifacts, acquisition sites, demographic characteristics, patient postures, and medical devices. These domain gaps can lead to undetected overfitting, making models incapable of generalization and, ultimately, unsuitable for clinical applications.
	
	2. \textbf{Generalization remains a key challenge for CXR applications.}
	Medical models can be evaluated and deployed in either in-distribution (ID) or out-of-distribution (OOD) settings. While these models often demonstrate excellent performance in ID settings, they frequently fail to maintain this level of expertise in OOD settings. Consequently, the AP and PP domain gaps raise concerns about the generalization ability of CXR models. Khorram et al. \cite{IGOS++} found that models tend to overfit to extraneous features such as singleton characters printed on CXR images. López-Cabrera et al. \cite{blackLung} discovered that even when the lung regions were replaced with black squares, many models still achieved an accuracy greater than 95\%. Therefore, rigorous evaluation of medical AI models necessitates assessing their performance in OOD settings to avoid "under-specification," which can lead to unanticipated poor performance during clinical deployment \cite{oodNeed}. These findings underscore the importance of testing the generalization ability of models, emphasizing that evaluations should consider not only ID performance but also OOD performance and attention maps.
	
	3. \textbf{Self-supervision for data-efficient transfer-learning and robust medical models.}
	Due to the limited availability of pediatric datasets and the time-consuming nature of annotating CXR images, self-supervised methods \cite{ssDomainGap, metateacherAAGap} are essential for generating robust pediatric CXR models with limited data. Traditional transfer learning typically requires a large number of annotated images to retrain the model for new domains. However, this process must be repeated for each new distribution shift, such as the introduction of new imaging equipment or deployment in a new clinic \cite{prepareMedicalImages}. This requirement for constant retraining and annotating significantly prolongs the lifecycle of medical imaging AI development and deployment, presenting a major barrier to their widespread adoption. Taeyoung et al. \cite{mae_UCSD} successfully applied Masked Autoencoders (MAE) \cite{mae_medical} to transfer models from adult to pediatric datasets. Although MAE facilitates the generalization of the feature encoder from adult to one pediatric dataset, the robustness of the model on other unseen pediatric datasets was not evaluated. Similarly, Azizi et al. \cite{simCLR_medical} utilized a representation learning strategy, SimCLR \cite{simCLR}, to effectively transfer models trained on natural images to medical images. While SimCLR can adapt the feature encoder to new domains, it requires hundreds of thousands of unlabeled images during the self-supervised contrastive learning stage.
	
	4.\textbf{Summary of key contributions.}
	\begin{itemize}
		\item \textbf{Lightweight U-Net model for Deep Contrast Enhancement (DCE)}: Recognizing that datasets from different sources can contain hospital-specific patterns and that models can easily overfit to noise such as corner text \cite{children2, children3, bias}, we proposed a lightweight U-Net model to perform DCE in a self-supervised manner. This approach highlights details in the lung area and suppresses other regions. By focusing on the lung area, DCE helps reduce domain gaps and results in robust CXR models with high performance in OOD settings.
		\item \textbf{Evaluating OOD performance of self-supervised methods}: In the task of transferring adult models to pediatric datasets with limited data, we evaluated the OOD performance of two self-supervised methods, MAE \cite{mae_medical} and SimCLR \cite{simCLR, simCLR_medical}. We found that high in-distribution performance on one pediatric dataset does not guarantee robust performance in OOD settings.
		\item \textbf{Self-supervised transfer learning framework (SCC)}: To efficiently build a generalizable pediatric CXR model, we presented a self-supervised transfer learning framework. This framework produces a robust pediatric model with superior generalization ability, requiring 10 times fewer images compared to the traditional transfer learning process.
	\end{itemize}	
	
	\section{Materials}
	
	To obtain a robust model under the constraint of limited pediatric CXR dataset size, transferring from large adult CXR datasets is required. Our model is built from TorchXRayVision \cite{xrv}, trained on 13 public adult datasets, including 731,075 images and 18 lung-related disease labels. We use three pediatric datasets, P1, P2 and P3, as shown in Table \ref{tab:dataset}. More dataset details are available in Appendix A.1.
	
	\begin{table}[!htbp]
		\caption{ \textbf{Dataset summary.} P1 is a private dataset. P2 is the PediCXR dataset \cite{PCXR}, a pediatric CXR dataset collected from a major pediatric hospital in Vietnam between 2020 and 2021. P3 \cite{UCSD_data} is the Guangzhou Women and Children's Medical Center (GWCMC) dataset, also known as the Kermany dataset.}
		\label{tab:dataset}
		\centering
		\begin{tabular}{llll}
			\toprule
			Dataset & Positive & Normal & Ages (years) \\
			
			\midrule
			P1 & 2824 & 2817 & 0-16\\
			P2  & 872 & 4365 & 0-12 \\   
			P3 & 1493  & 1583 & 1-5 \\
			\bottomrule 
		\end{tabular}	
	\end{table}

	\begin{figure}[!hbpt]
		\centering  
		\includegraphics[width=0.75\linewidth]{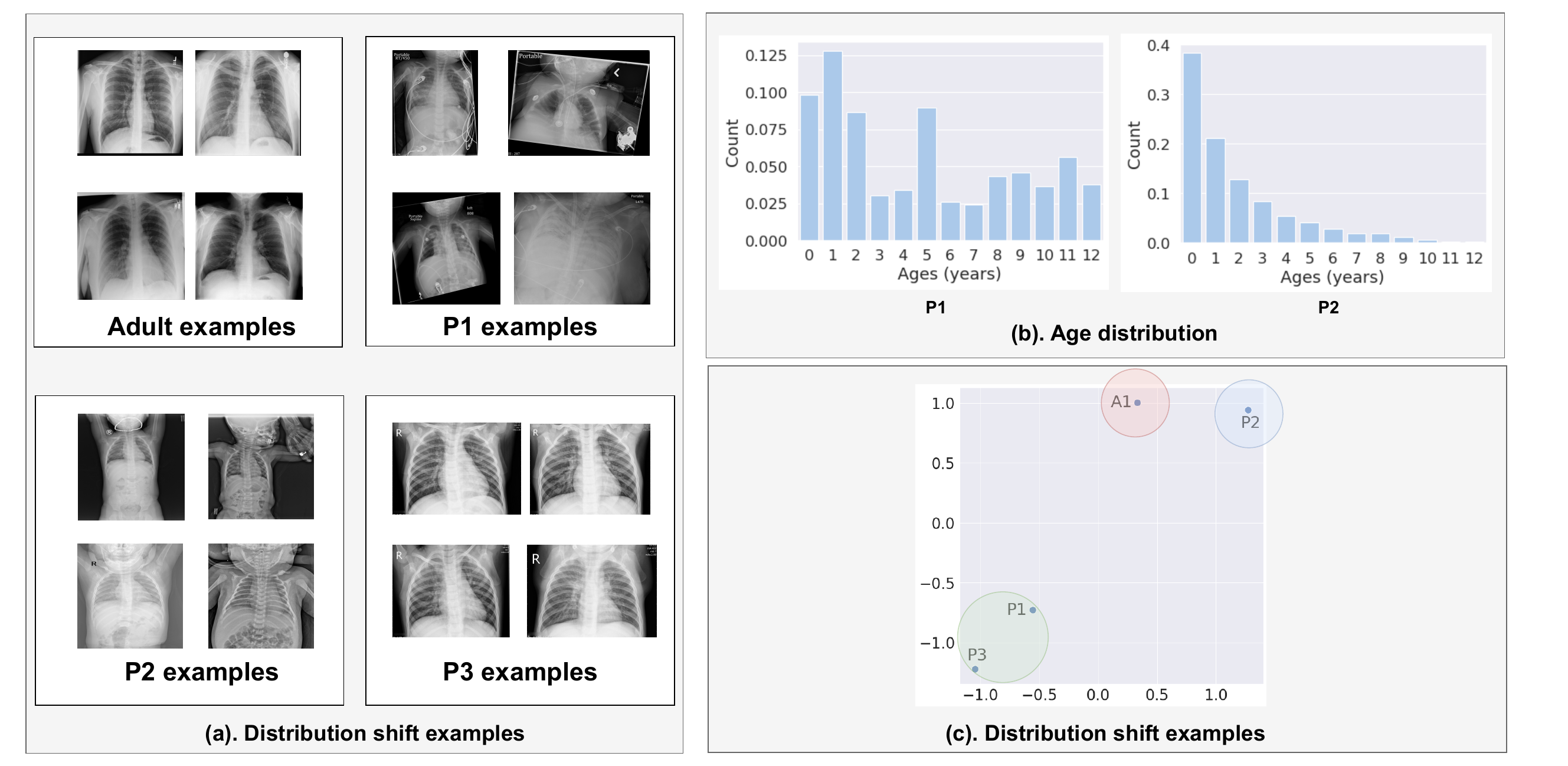}
		\caption{\textbf{Dataset description.} (a) Image examples. CXR images from different sources exhibit different attributes, implying the aforementioned two domain gaps: AP and PP domain gap. (b) Age distribution of P1 and P2 datasets. While P2 mainly comprises pediatrics under 2 years old, P1 has a relatively more uniform age distribution over 0-12 years old. (c)Domain gaps among pediatric and adult datasets. X and Y axes are the embedding spaces after we applied multidimensional scaling on the distribution distance matrix. When P1 and p3 are relatively closer to each other, the rest of the datasets all have a huge domain gap with others. It's worth noting that for P1, the distance to P2 is even further than the distance to A1, which emphasizes that we should consider not only the AP domain gap but also the PP domain gap.} 
		\label{fig:data_desc}  
	\end{figure}
	
	As shown in Figure \ref{fig:data_desc}(a), the CXR images from different sources exhibit different attributes, implying the aforementioned two domain gaps: AP and PP domain gap. Figure \ref{fig:data_desc}(b) demonstrates the age distribution of P1 and P2 while we can't find the age information of P3. It shows that while P2 mainly comprises pediatrics under 2 years old, P1 has a relatively more uniform age distribution over 0-12 years old, revealing the domain gap among different populations. We then measured the domain gap \cite{domainGap_quantify, domain_measure3} among the mentioned three pediatric datasets and one adult dataset \cite{KaggleRD_data1,kaggleRD_data2}, A1. The result is shown in Figure \ref{fig:data_desc}(c). More details about the quantification method are described in Appendix A.2. Figure \ref{fig:data_desc}(c) implies that P1 and P3 are relatively closer to each other, and the rest of the datasets all have a huge domain gap with others. It's worth noting that for P1, the distance to P2 is even further than the distance to A1, which emphasizes that we should consider not only the AP domain gap but also the PP domain gap.

	\section{Framework for Robust Pediatric CXR Models}

	\begin{figure*}[!b]
		\centering  
		\includegraphics[width=0.85\linewidth]{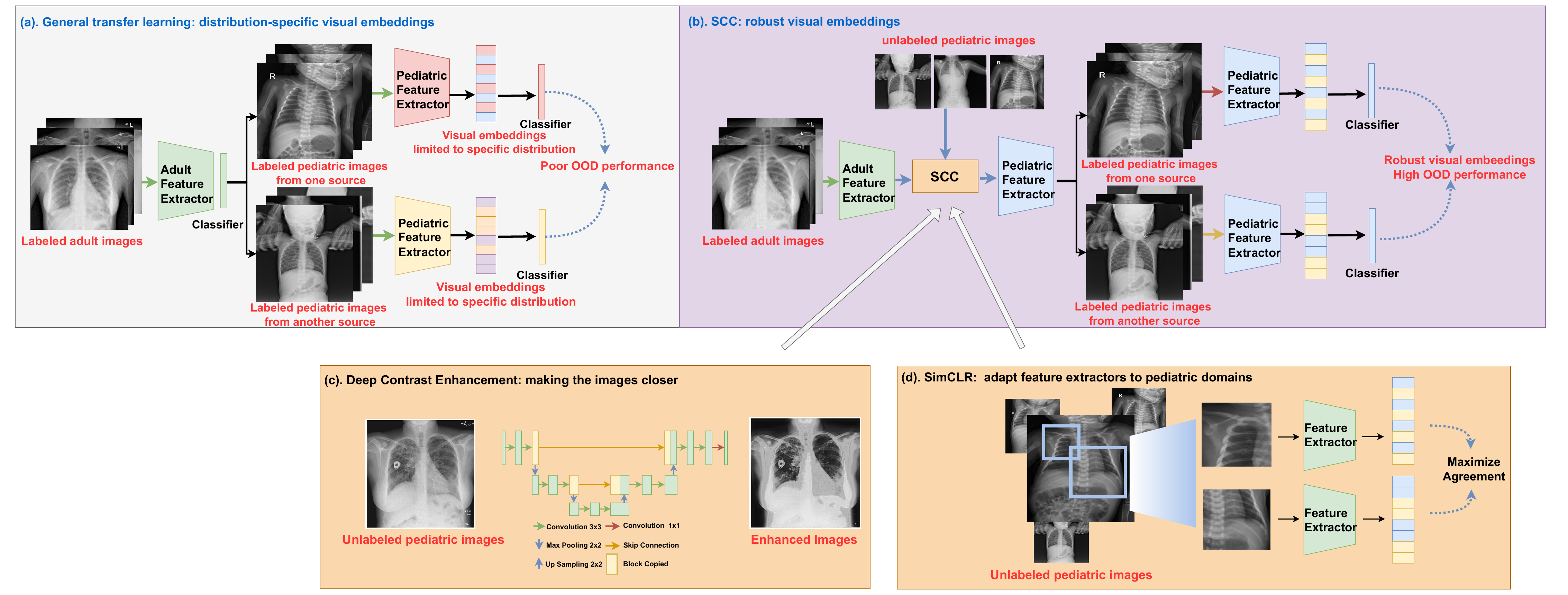}
		\caption{\textbf{Architectures of the framework SCC} (a) represents the traditional transfer learning process, which directly retrains the pre-trained adult model on pediatric images. This approach can lead to model overfitting to hospital or population-specific biases, resulting in poor generalization ability and unsuitability for clinical settings. (b) illustrates the proposed SCC framework, which integrates two self-supervised approaches: (c) making images more similar and (d) adapting the feature encoder to pediatric domains to overcome the AP and PP domain gaps. Consequently, SCC can build generalizable pediatric models with high OOD performance.} 
		\label{fig:overall_arch}  
	\end{figure*}
	
	The goal of our work is to build a robust pediatric model that can maintain good classification ability on unseen datasets. To achieve this, we must overcome two domain gaps: AP and PP domain gaps. However, due to the limited size of the pediatric CXR dataset, transferring models from adult CXR images to pediatric CXR images is challenging and can result in unstable models with low generalization ability.
	
	Generally, the influence caused by the domain gap during transfer learning can be alleviated by the following two methods: First, making the CXR images more similar on the input end, such as using lung segmentation \cite{deepCovidXR}; second, adapting the feature encoder to new domains on the feature end, such as with test-time training \cite{testTimeTraining} and SimCLR \cite{simCLR}.
	
	We propose a framework that combines transfer learning with Self-supervised Contrastive learning, augmented by an unsupervised Contrast enhancement technique (SCC), as shown in Figure\ref{fig:overall_arch}. Deep Contrast Enhancement is applied to the input end, while contrastive learning is utilized for the feature end.
	
	\subsection{Making the images closer: Pixel-Level deep contrast enhancement}
	Our purpose is to enhance the contrast within the lung area while suppressing other regions in a self-supervised manner. This approach ensures that the subsequent classification model focuses on the lung regions rather than other areas. Consequently, the generalization ability of the classification model is further improved, as the lesion parts should exhibit similar characteristics despite various noises. Generally, the proposed DCE can function as a free and lightweight lung segmentation tool and can be integrated into any standard medical image processing pipeline.
	
\begin{figure*}[!t]
		\centering  
		\includegraphics[width=0.85\linewidth]{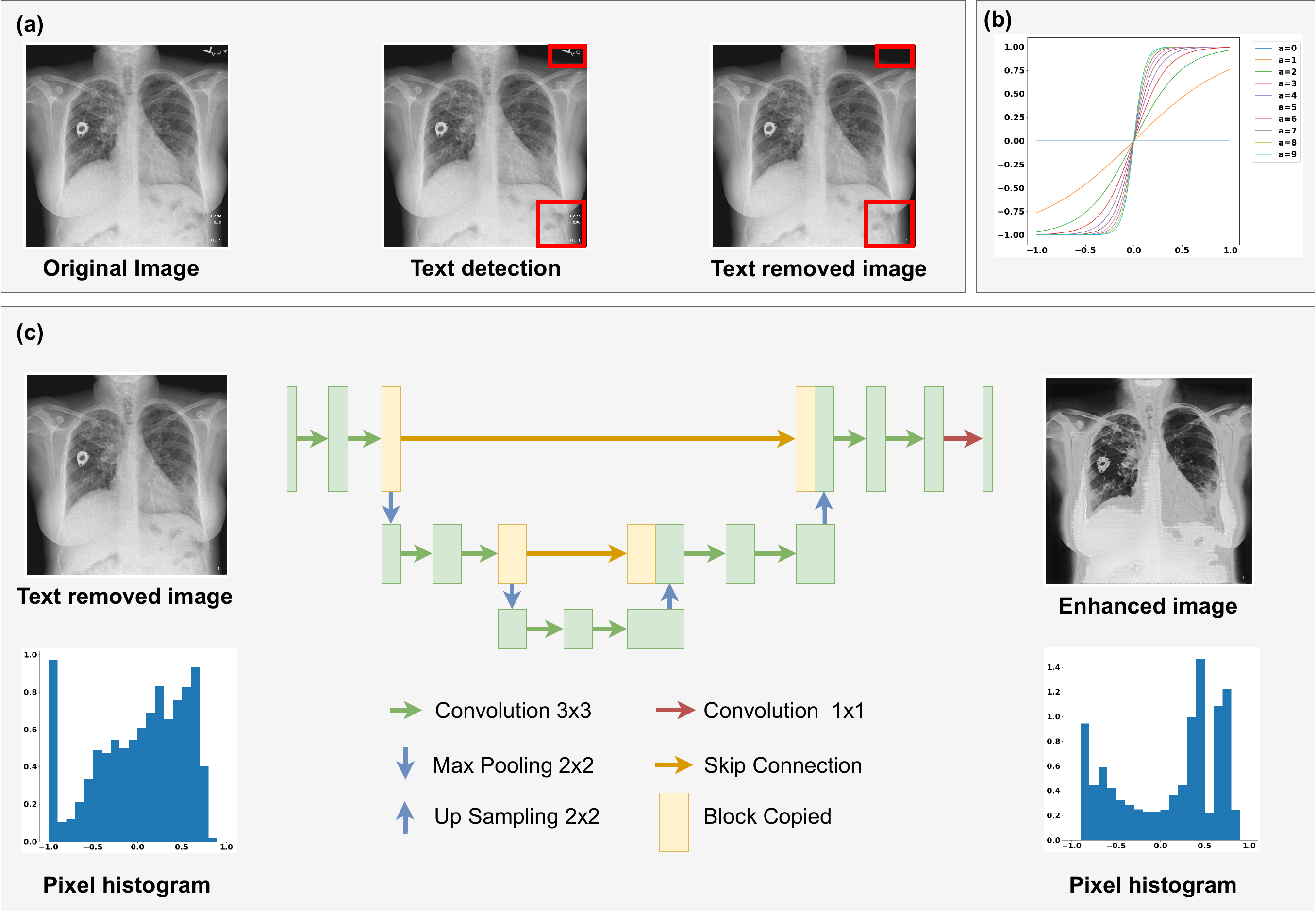}
		\caption{\textbf{Overview of the DCE.} (a) The first stage: Text removal. (b) Lung enhancement curve. When operating on a pixel-wise basis, this plot shows that a greater learnable parameter $\alpha$ can map the original pixels to a wider dynamic range, while a smaller $\alpha$ offers a narrower range. Therefore, with a suitable $\alpha$, the details of the lung area can be highlighted, and other regions can be suppressed, generating a clearer version of the images for the subsequent classification model. (c) DCE architecture and examples. Comparing the images and the pixel histograms, pixels of the original images are concentrated in the middle range, which blurs the lesion pixels with surrounding objects like ribs or the lung background. Conversely, the enhanced image has a more uniform pixel intensity distribution, which helps highlight the lesion pixels in the lung area and suppresses other regions like the abdomen.} 
		\label{fig:dce_flow}  
	\end{figure*}

	As shown in Figure \ref{fig:dce_flow}, DCE is a two-stage image processing method: 1. Text removal and 2. Contrast enhancement. Figure \ref{fig:dce_flow}(a) details the text removal stage. We first use EasyOCR \cite{easyOCR} to detect the text location and then apply DeepFill \cite{deepfill} for inpainting. In this stage, only text pixels are modified; thus, the image quality is preserved, ensuring no loss of relevant information.

	The details of the second stage, contrast enhancement, are described as follows.
	
	\textbf{Lung Enhancement Cure (LE-curve)} After normalizing all pixels to [-1,1], we use $tanh()$ curve to map an original CXR image to its enhanced version in pixel-wise, as shown in Eq.(1).
	\begin{equation}
		LE(I(x);\alpha)=y\_shift+tanh(\alpha \times (I(x)+x\_shift)),
	\end{equation}
	where $LE(I(x); \alpha)$ is the enhanced pixel value of the given input pixel value $I(x)$, and $\alpha$ is the learnable parameter that adjusts the sensitivity of the LE-curve, providing the flexibility to operate on each pixel.
	
	An illustration of the LE-curve with different learnable parameters $\alpha$ is shown in Figure \ref{fig:dce_flow}(b). It is evident that with a greater $\alpha$, the LE-curve applies a wider dynamic range to input pixels near 0 and a narrower dynamic range to input pixels near -1 or 1. This capability is conducive to highlighting the input pixels near the middle value and suppressing the input pixels near the ends, thereby providing a lung-enhanced image, as shown in Figure \ref{fig:dce_flow}(c).
	
	\textbf{Architecture} To find the best-enhanced image, a transformation matrix (A), composed of the learnable parameter $\alpha$ corresponding to each pixel of the original image, is required. This matrix has the same dimensions as the input image. We modify the U-Net \cite{unet} architecture and propose a shallow U-Net to generate A for each input image, as shown in Figure \ref{fig:dce_flow}(c). The input to the shallow U-Net is a grayscale CXR image, and the output is the corresponding transformation matrix, A. We can then obtain the lung-enhanced image by performing element-wise multiplication of the original image and the transformation matrix, A.

	\textbf{Loss functions.} To preserve the original information in the images while achieving self-supervision of the lung enhancement preprocessing, we propose a series of loss functions.
	\begin{enumerate}
		\item \textbf{Adaptive loss.} This loss function encourages the learnable parameter $\alpha$ to follow our principle, where $\alpha$ provides the lung area with a wider dynamic range and a narrower range for other regions. It can be expressed as follows:
		\begin{equation}
			L_{adapt}=\frac{||A-TS \times G(X;\sigma, \theta)||_2^2}{2\times B \times C \times W \times H},
		\end{equation}
		
		\begin{equation}
			\sigma=Sort(Kmeans(X,5))[3],
		\end{equation}
		
		\begin{equation}
			\begin{split}
				\theta=Min(&Sort(Kmeans(X,5))[2], \\
				&Sort(Kmeans(X,5))[4]),						
			\end{split}
		\end{equation}
		
		Where A is the transformation matrix, TS is the transformation strength and $G(\cdot)$ represents the Gaussian function. X is the input image matrix, while B, C, W, and H denote batch size, channel number, image width and image height, respectively. For the Gaussian function, we first apply 5 clusters of K-means to the original images and sort the cluster centers. Then, we choose the third cluster center as the mean and the minimum value between the second and the fourth cluster centers as the variance. The assumption is that each CXR image comprises five parts, in ascending order of pixel intensity: image background, lung background, lesions, soft tissues, and bones.

		\item \textbf{Local conflict loss}. To preserve the contrast information among pixels, we add a local conflict loss function, which is defined as follows:
		\begin{equation}
			L_{lcl}=\frac{\sum_{i}\sum_{j \in Region(i)}(Y_i>Y_j xor X_i>X_j)}{4 \times B \times C \times W \times H},
		\end{equation}
		where $i$ denotes the position of one pixel, and $Region(i)$ denotes the position of four neighboring pixels (top, down, left, right) of pixel $i$. $Y$ and $X$ represent the enhanced pixel values and the original pixel values, respectively. 
		
		\item \textbf{Region conflict loss}. Additionally, we introduce the region conflict loss function aimed at preserving contrast information among regions. Initially, we flatten each $k \times k$ square within both the original and enhanced images. Subsequently, we compute the discrepancy between the gram matrices of these flattened matrices. Drawing inspiration from artistic style transfer techniques \cite{styleTransfer}, the flattening process negates spatial information, thus facilitating the preservation of contrast information from a broader perspective. The formulation of the region conflict loss function is as follows:
		
		\begin{equation}
			L_{rcl}=\frac{||G_{FX}-F_{FY}||_2^2}{4N^2M^2},
		\end{equation}
		\begin{equation}
			FX=SquareFlatten(X; kernel\_size),
		\end{equation}
		\begin{equation}
			FY=SquareFlatten(Y; kernel\_size),
		\end{equation}
		where $G_X \in R^{T \times T}$ is the gram matrix of $X$ which has $T$ rows. $FX$ and $FY$ present the flattened matrices of X and Y, respectively. $N$ and $M$ are the width of $FX$ and $FY$. $SquareFlatten(X; kernel\_size)$ is a function to flatten each $kernel\_size$ square of matrix $X$.
	\end{enumerate}
	
	As depicted in Figure \ref{fig:dce_flow}(c), the histograms of original images and enhanced images suggest that pixels in the original images tend to cluster within the middle range. This clustering effect often leads to the blurring of lesion pixels with surrounding objects such as ribs or the lung background. Conversely, the enhanced image exhibits a more evenly distributed pixel intensity, which aids in highlighting lesion pixels within the lung area while simultaneously dampening the contrast in other regions, such as the abdomen.
	
	\subsection{Making the feature closer: SimCLR}  
	Since previous studies have demonstrated the effectiveness of pretraining on extensive unlabeled datasets in mitigating distribution shift issues, our focus lies in harmonizing the feature encoder to accommodate both the pretrained adult domain and target pediatric domain, particularly in scenarios with limited dataset sizes. This alignment ensures that similar features are inputted into the subsequent classification layer.
	
	With a well-trained adult CXR model, we fine-tune the feature encoder $f({\cdot})$ in a self-supervised manner on unlabeled pediatric datasets to produce a robust visual embedding by minimizing the contrastive loss function \cite{simCLR}, as shown in Eq.(9). 
	\begin{equation}
		l_{i,j}=-log\frac{exp(sim({z_i, z_j})/\tau)}{\sum_{k=1}^{2N}\mathbf{1}_{[k\neq i]}exp(sim(z_i, z_k)/\tau)}
	\end{equation}
	where $i$, $j$ stand for the two views coming from the same image, respectively. $sim(\cdot,\cdot)$ is cosine similarity between two vectors, and $\tau$ is a scalar denoting the temperature.
	
	Specifically, SimCLR learns embeddings by distinguishing whether the output features originate from various augmented views of the same training example. In a batch of images, each image $X_i$ generates two views with distinct augmentations, denoted as $x_{2k-1}$ and $x_{2k}$. These two images undergo mapping via a pre-trained feature encoder and a non-linear transformation head, yielding visual embeddings $z_{2k-1}$ and $z_{2k}$, which are leveraged for computing the contrastive loss objective. Following this phase of intermediate self-supervised training, the transformation head is discarded, and the feature encoder is utilized for subsequent supervised training.

	\section{Experiments and Results}
	
	\subsection{Experiment setting}
	To ensure the robustness our method, we compared SCC to both supervised models and self-supervised methods using transfer learning. We choose TorchXRayVision\cite{xrv} as the baseline for supervised models, SimCLR\cite{simCLR_medical} and MAE\cite{mae_UCSD, mae_medical} as baselines for self-supervised methods. To test the generalization ability of our classification models, we use the P1 dataset as the in-distribution dataset and p2, p3 as the out-of-distribution datasets which reflects a variety of realistic distribution shifts due to data acquisition devices, clinical demographics and so on, as shown in Figure \ref{fig:data_desc}.

	1. In-distribution performance evaluation: The model is trained and test on the in-distribution dataset, P1, under 5-fold cross-validation settings. 
	
	2. Zero-shot out-of-distribution performance evaluation: The out-of-distribution datasets, P2 and P3, are split into $D_{out}^{train}$ and $D_{out}^{test}$ with the ratio 9:1. The model is evaluated on $D_{out}^{test}$ without any further fine-tuning using OOD data.
	
	3. Few-shot fine-tuning and performance evaluation: The model is further fine-tuned using some fraction of $D_{out}^{train}$ and then test on out-of-distribution test samples $D_{out}^{test}$.

	Each experiment is under 5-fold cross-validation, and we tried both linear probing and whole network fine-tuning to achieve the best performance. Further hyper-parameter settings are available in Appendix B.

	\subsection{Performance evaluation}
	\begin{table}[!htbp]
		\caption{\label{tab:performance} \textbf{Performances.} Here we reported the AUC of experiments based on the three pediatric datasets and the star (*) stands for our proposed method. SCC has the highest zero-shot and few-shot AUC scores, which implies that it can help build robust and generalizable pediatric models under limited datasets when transferring from large adult models.}
		\centering
		\begin{tabular}{lllll}
			\toprule
			
			Dataset & Method & In-distribution & OOD(0\%) & OOD(100\%)\\
			
			\midrule
			\multirow{6}{*}{P1} & Xrv & 0.89  $\pm$ 0.01& \multirow{6}{*}{\textbackslash}	& \multirow{6}{*}{\textbackslash}  \\
			& SimCLR			& 0.91 $\pm$ 0.02 & 									&  \\
			& MAE			& \textbf{0.92 $\pm$ 0.01} & 									&  \\
			& DCE+MAE* 			& 0.9 $\pm$ 0.02 &  									&  \\ 			
			& DCE* 			& 0.91 $\pm$ 0.01 &  									&  \\ 
			& SCC*	& \textbf{0.92 $\pm$ 0.01} &  									&  \\
			\midrule 
			\multirow{6}{*}{P2} & Xrv & \multirow{6}{*}{\textbackslash} & 0.52 $\pm$ 0.03 & 0.75  $\pm$ 0.03\\
			& SimCLR			&								  		& 0.60 $\pm$ 0.02 & 0.85  $\pm$ 0.02\\
			& MAE				&								  		& 0.43 $\pm$ 0.01 & 0.78 $\pm$ 0.02 \\			
			& DCE+MAE*			&										& 0.44 $\pm$ 0.03 & 0.89 $\pm$ 0.04 \\ 			
			& DCE*				&										& 0.67 $\pm$ 0.01 & 0.83  $\pm$ 0.01\\ 
			& SCC*			&								  		& \textbf{0.70 $\pm$ 0.01} & \textbf{0.90 $\pm$ 0.01} \\
			\midrule 
			\multirow{6}{*}{P3} & Xrv & \multirow{6}{*}{\textbackslash} & 0.81  $\pm$ 0.04& 0.95 $\pm$ 0.01  \\
			& SimCLR			&	 								  	& 0.89 $\pm$ 0.03 & 0.99  $\pm$ 0.01\\
			& MAE				&	 								  	& 0.85 $\pm$ 0.01 & 0.99 $\pm$ 0.01 \\
			& DCE+MAE*				&										& 0.83 $\pm$ 0.02 & 0.99  $\pm$ 0.01\\						
			& DCE* 				&	 								  	& 0.91  $\pm$ 0.01& 0.99 $\pm$ 0.01 \\ 
			& SCC*				&										& \textbf{0.92 $\pm$ 0.02} & 0.99 $\pm$ 0.01 \\
			\bottomrule 
		\end{tabular}	
	\end{table}
	
	\begin{figure}[!htbp]
		\centering  
		\includegraphics[width=0.85\linewidth]{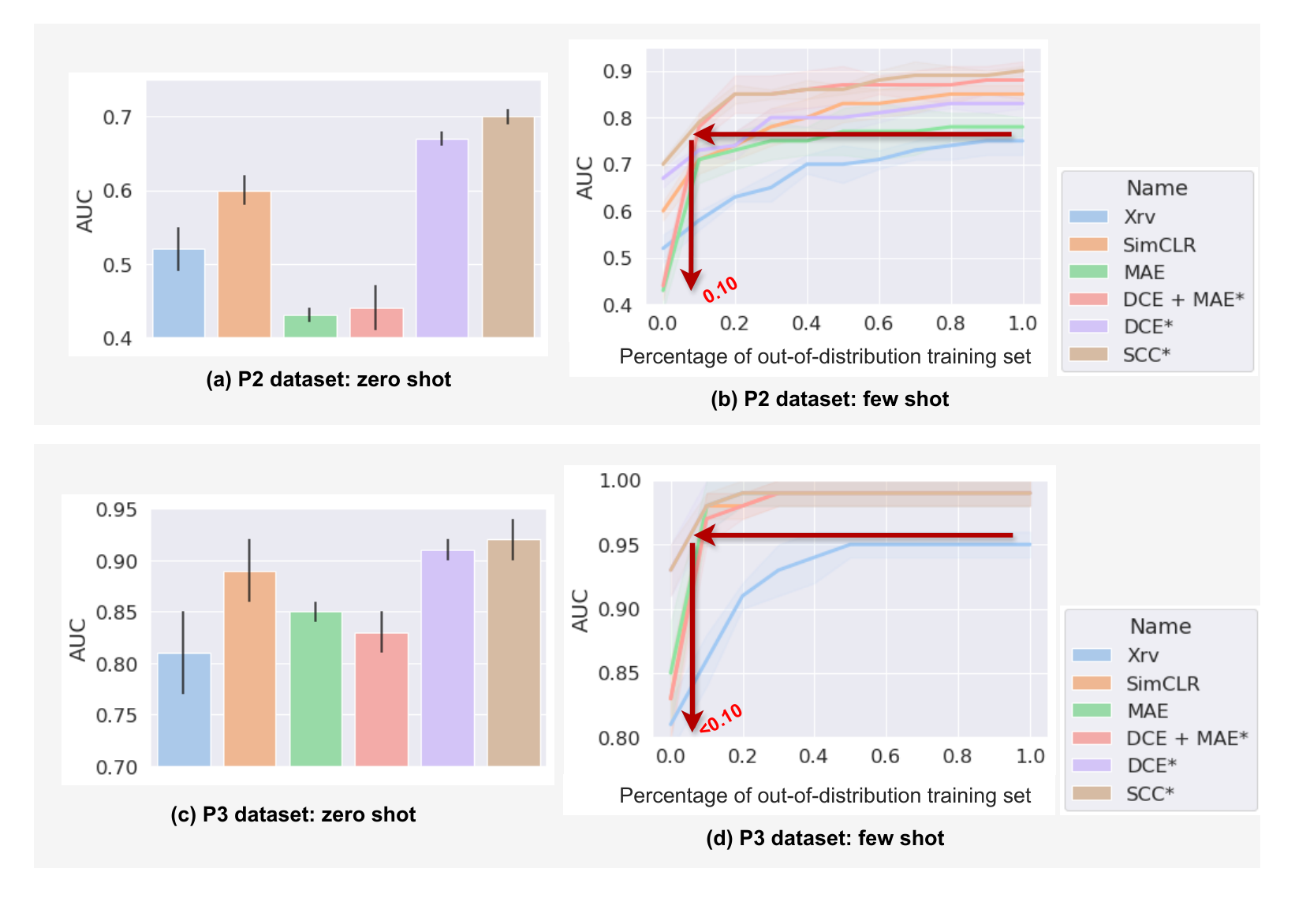}
		\caption{\textbf{OOD performances}. Overview of the OOD performances, demonstrating the high generalization ability of SCC alongside strong baselines. Figures (a) and (c) depict the zero-shot performance, indicating SCC's superior OOD classification performance even without access to retraining data in a new clinical setting. Figures (b) and (d) display the few-shot learning performance with varying training ratios of the P2 and P3 datasets, respectively. The best transfer learning performance of the supervised baseline is matched by SCC with access to less than 10\% of the labeled images of P2 and P3 datasets, which indicates that our proposed framework can achieve comparable accuracy as baseline specialized models using 10 times less labeled data.} 
		\label{fig:fewshot_performance} 
	\end{figure}
	
	\begin{figure}[!htbp]
		\centering  
		\includegraphics[width=0.85\linewidth]{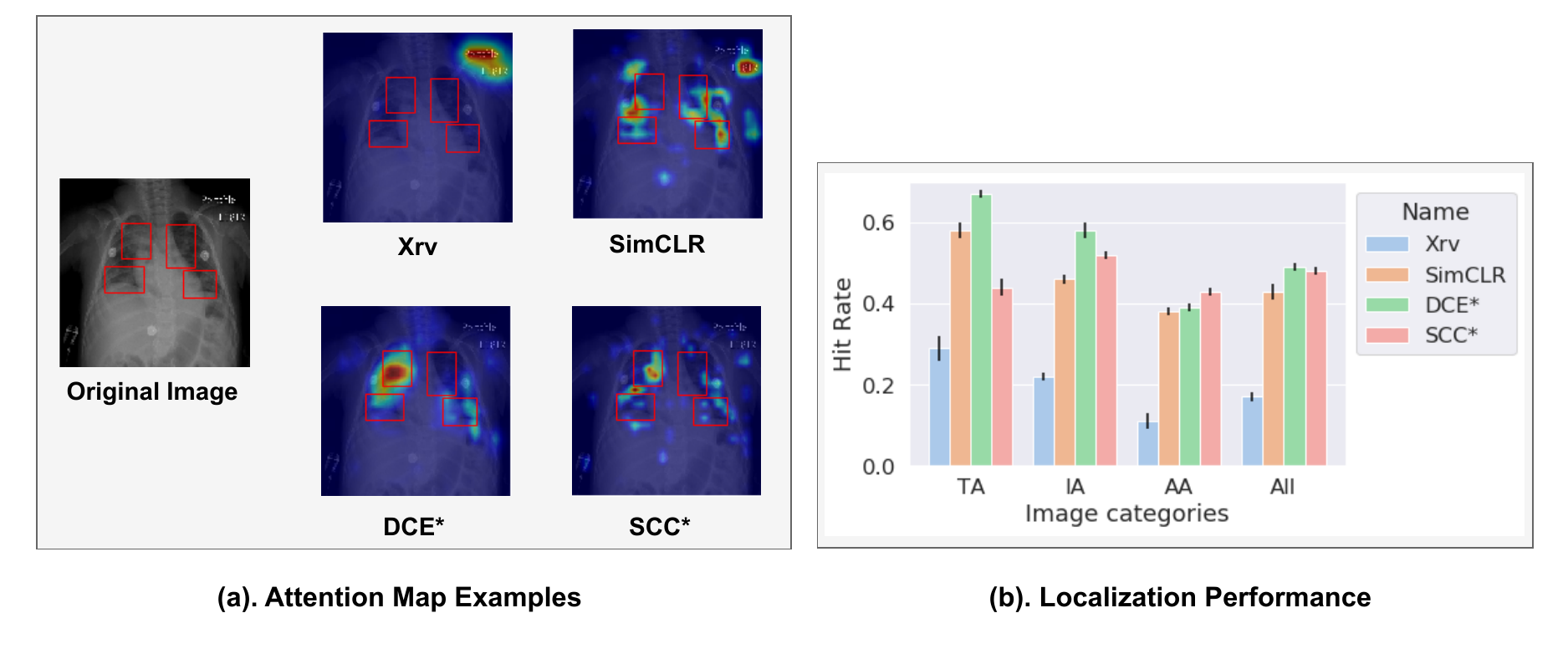}
		\caption{\textbf{Localization performance.} Figure (a) shows an positive example and the corresponding attention maps. The red bounding box is the lesion part, drawn by an expert radiologist. Notably, both Xrv and SimCLR appear to be influenced by text and noise, hindering their generalization ability. Conversely, DCE and SCC yields more precise attention maps, concentrating inside the lung area. Figure (b) is the quantification scores of hit rate\cite{hitrate}. The positive images of the P1 dataset contains three categories: typical appearance (TA), Indeterminate Appearance (IA), and Atypical Appearance (AA). "All" means the average scores of all the images. Both DCE and SCC exhibit significant improvement compared with the strong supervised baseline, which suggests the robustness and high generalization ability of our proposed framework. Though SimCLR achieves relateively higher scores on the TA type, it shows lower scores for other types, potentially due to inherited biases from the pretrained adult model.
		} 
		\label{fig:attentionMap}  
	\end{figure}

	\textbf{SCC leads to statistically significantly improved generalization ability.}  
	Figure \ref{fig:fewshot_performance} and Table \ref{tab:performance} provide an overview of the OOD performances, demonstrating the high generalization ability of SCC alongside strong baselines. SCC achieves superior OOD classification performance while significantly reducing the need for labeled data. Notably, it enhances OOD performance in terms of AUC from 0.52 to 0.70 on P2 and from 0.81 to 0.92 on P3, even without access to retraining data in a new clinical setting.
	
	\textbf{SCC requires fewer labeled images to get comparable performance compared with supervised models on OOD settings.} As shown in \ref{fig:fewshot_performance}(c) and (d), the best transfer learning performance of the supervised baseline is matched by SCC with access to less than 10\% of the labeled images of P2 and P3 datasets, which indicates that our proposed framework can achieve comparable accuracy as baseline specialized models using 10 times less labeled data.
	
	\textbf{SCC demonstrates superior localization performance.} We evaluate attention maps of the P1 dataset with iGOS++\cite{IGOS++} to assess the localization performance, as shown in Figure \ref{fig:attentionMap}. Given MAE's poor generalization ability, we exclude it from our analysis. Figure \ref{fig:attentionMap}(a) displays the original image while the red bounding box was drawn by an expert radiologist. Notably, the baseline model appears to be influenced by text and noise, hindering its generalization ability. Conversely, DCE and SCC yields more precise attention maps, concentrating inside the lung area. Moreover, we use the hit rate\cite{hitrate} to quantify the localization performance of the attention maps, as shown in Figure \ref{fig:attentionMap}(b). The hit rate is based on the pointing game set-up, in which credit is given if the most representative point identified by the visualization method lies within the ground-truth segmentation. The positive images of the P1 dataset are further split into three categories: typical appearance (TA), Indeterminate Appearance (IA), and Atypical Appearance (AA). Both DCE and SCC exhibits significant improvement compared with the strong baseline, which suggests the robustness and high generalization ability of our proposed framework. Though SimCLR achieves relateively higher scores on the TA type, it shows lower scores for other types, potentially due to inherited biases from the pretrained adult model.

	\subsection{Framework Generalization Ability}
	To test the generalization ability of SCC, we also run it on three benchmark breast ultrasound datasets: B1, B2, and B3. We use B1\cite{b1} as the ID training dataset, while B2\cite{b2} and B3\cite{b3} served as the OOD test datasets. Transferred from ResNet50 pretrained on the ImageNet-1K dataset, SCC successfully improves the zero-shot performance from 0.84 to 0.87 on B2 and from 0.73 to 0.77 on B3. After fine-tuning on B2 and B3, the AUC increased from 0.94 to 0.95 on B2 and from 0.78 to 0.83 on B3. These results suggest that SCC can function as an insertable framework to help build robust models for medical images. More details about the breast ultrasound datasets are available in Appendix C.

	\section{Discussion}
	
	This study introduces a self-supervised transferring framework that effectively transfers adult CXR models to pediatric datasets, demonstrating strong robustness and high performance on previously unseen datasets. Our main takeaways are as follows: (a) A lightweight self-supervised U-Net model (DCE) that can enhance the contrast within the lung area while suppressing other regions, reducing the impact of image variations and producing high-quality embedding across diverse pediatric CXR images coming from different sources.(b) By integrating SimCLR with DCE, we introduce a self-supervised transferring framework, which achieves superior performance in both ID and OOD settings. It requires 10 times fewer labeled images to match up with the best performance of traditional supervised transfer-learning settings. Our observations indicate significant improvements in generalization ability when transferring from adult to pediatric CXR images and from natural images to breast ultrasound images. However, it is important to note that our task focused mainly on pediatric CXR images related to viral pneumonia and breast ultrasound images related to malignant lesions. Our work did not undergo rigorous clinical testing and therefore cannot be used in clinical practice. We hope our work contributes to the AI-based medical diagnosis domain and accelerates relevant model development. We plan to explore our methods on multi-label tasks, including other lung-related diseases, and leverage both radiology reports and CXR images to develop more explainable and generalizable multi-modal pediatric CXR models.
	
	\bibliographystyle{unsrtnat}
	\bibliography{references}
	
	
	\appendix
	
	\section{Appendix: Pediatric CXR datasets}
	
	\subsection{Dataset description}
	We used the following three pediatric datasets. P1 is a private dataset including 5641 CXR images from children between 0-16 years old. This dataset has two labels: COVID-19 and normal. P2 is the PediCXR dataset\cite{PCXR}, a pediatric CXR dataset of 9125 studies retrospectively collected from a major pediatric hospital in Vietnam between 2020 and 2021. Each scan was manually annotated by a pediatric radiologist with more than ten years of experience. The dataset was labeled for 36 critical findings and 15 diseases. We composed a subset by mixing the images with labels of pneumonia, pleuro-pneumonia, broncho-pneumonia and normal. P3\cite{UCSD_data} is the Guangzhou Women and Children's Medical Center (GWCMC) dataset, also known as the Kermany dataset. This dataset comprises 5,856 anteroposterior (AP) chest radiographs from children ages 1–5. The dataset includes three labels: normal, bacterial pneumonia, or viral pneumonia, including 5,232 and 624 training and test samples, respectively. Two physicians labeled all images, with a third physician verifying all test dataset labels. We used the images with labels as normal and viral pneumonia. The summary of datasets is shown in Table \ref{tab:dataset}. The age distribution of P1 and P2 is shown in Figure \ref{fig:data_desc} while we couldn't find the age information on the official website of P3.
	
	\subsection{Domain gap measurement}
	To quantify the domain gaps of multiple datasets, we need the domain metrics to be domain-agnostic so that the gaps between	different datasets can be more meaningful and comparable. We use pixel intensities and texture features to evaluate the feature distributions of different datasets. More specifically, the gray-level pixels' mean value and standard deviation are considered as the color features. Gray-Level Co-occurrence Matrix features \cite{domain_measure3} (Angular Second Moment, Homogeneity, Contrast, Correlation) of 4 directions are adopted as the texture features. Given the above feature distributions, the domain gap is measured as the Maximum Mean Discrepancy between each dataset. We then apply multidimensional scaling to the distance matrix and get the 2-dimensional domain gap distance plot.

	\section{Appendix: Hyper-parameters}
	
	We deployed Ray-Tune\cite{rayTune} to realize careful hyper-parameter tuning. We use the Distributed Asynchronous Hyper-parameter Optimization algorithm\cite{hyperOPT} to do the hyper-parameter searching and the Adam\cite{Adam} optimizer with the initial learning rate picking in the range of $[1e^{-6},1e^{-3}]$ and weight decay of $[1e^{-5},1e^{-2}]$. We used the LambdaLR scheduler with the lambda in the range of $[0.6, 1]$, and the batch size was picked from $[32, 64, 128]$.
	
	During the training, we used an augmentation strategy consisting of random cropping with a ratio from 0.6 to 1.0, scaling to 224 $\times$ 224 pixels, random rotation up to $15^{\circ}$, and normalizing each pixel to the range of $[-1024, 1024]$ as required by TorchXRayVision\cite{xrv}. For the P2 dataset, as it contains images mostly coming from children under 2 years old and includes a large series of unseen noises like others' hands, we manually used rectangle masks to pick out the lung region part.
	
	\section{Appendix: Experiments on Breast Ultrasound Datasets}
	
	\begin{table}[!htbp]
		\caption{ \textbf{Dataset summary.} B1 was collected from 1064 female patients acquired via four ultrasound scanners during systematic studies at the National Institute of Cancer (Rio de Janeiro, Brazil). B2 collects at baseline including breast ultrasound images among 600 female patients in ages between 25 and 75 years old in 2018. B3 was collected by five radiologists at medical centers in Poland in 2019–2022. All images were manually annotated and labeled by radiologists via a purpose-built cloud-based system. }
		\label{tab:datasetBreast}
		\centering
		\begin{tabular}{lll}
			\toprule
			Dataset & Malignant & Benign \\
			
			\midrule
			B1 & 607 & 1268 \\
			B2  & 210 & 437  \\   
			B3 & 98  & 158 \\
			\bottomrule 
		\end{tabular}	
	\end{table}
	\subsection{Materials}
	To test the generalization ability of SCC, we also run it on three benchmark breast ultrasound datasets: B1, B2, and B3, as shown in Table \ref{tab:datasetBreast}. B1 comprises 1875 anonymized images from 1064 female patients acquired via four ultrasound scanners during systematic studies at the National Institute of Cancer (Rio de Janeiro, Brazil). The dataset includes biopsy-proven tumors divided into 722 benign and 342 malignant cases. B2 collects at baseline including breast ultrasound images among 600 female patients in ages between 25 and 75 years old in 2018.  It consists of 780 images, categorized into three classes: normal, benign, and malignant. To align with other datasets, we use only benign and malignant images. B3 consists of images of 154 benign tumors, 98 malignancies and 4 normal breasts. It was collected by five radiologists at medical centers in Poland in 2019–2022. All images were manually annotated and labeled by radiologists via a purpose-built cloud-based system.

	\begin{figure}[!htbp]
		\centering  
		\includegraphics[width=0.85\linewidth]{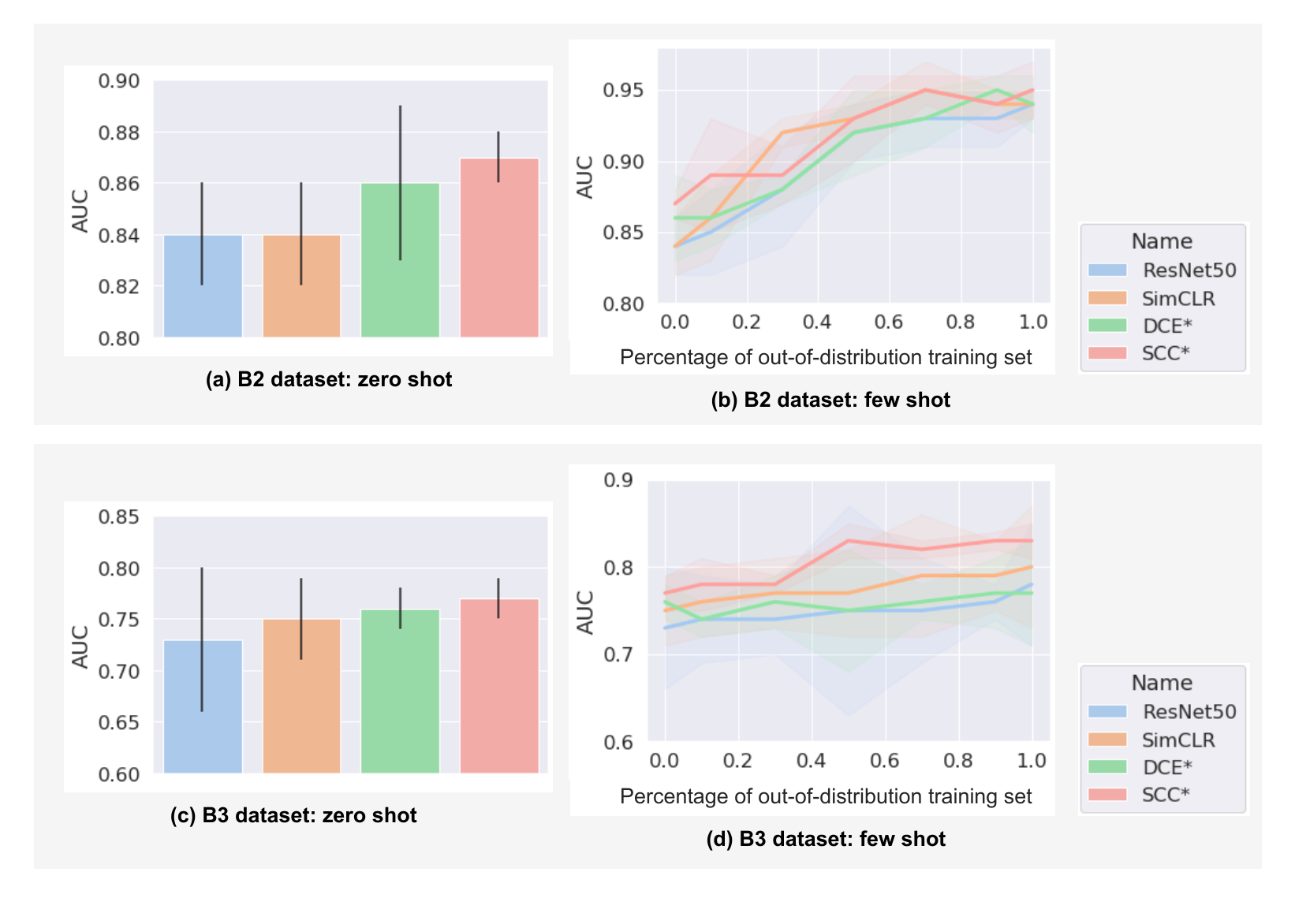}
		\caption{\textbf{OOD performances of the breast datasets}. Overview of the OOD performances, demonstrating the high generalization ability of SCC alongside strong baselines. Figures (a) and (c) depict the zero-shot performance, indicating SCC's superior OOD classification performance even without access to retraining data in a new clinical setting. Figures (b) and (d) display the few-shot learning performance with varying training ratios of the B2 and B3 datasets, respectively.} 
		\label{fig:fewshot_performance_Breast} 
	\end{figure}
	
	\subsection{Experiment settings}
	We use the ResNet50 model pretrained on ImageNet-1K dataset as our base model. We use B1\cite{b1} as the ID training dataset, while B2\cite{b2} and B3\cite{b3} served as the OOD test datasets. B2 and B3 are further split into $D_{out}^{train}$ and $D_{out}^{test}$ with the ratio 8:2. The model fine-tuned on B1 dataset is evaluated on $D_{out}^{test}$ of B2 and B3 for zero-shot evaluation. To assess the few-shot learning performance, the model will be further fine-tuned using some fraction of $D_{out}^{train}$ and then test on out-of-distribution test samples $D_{out}^{test}$. As for the hyper-parameter tuning, we use the same settings of the pediatric CXR task. We use two NVIDIA RTX A6000 graphics cards with 49140 Mib memory for each.
	
	\subsection{Performance evaluation}
	
	Figure \ref{fig:fewshot_performance_Breast} and Table \ref{tab:performanceBreast} provide an overview of the OOD performances, demonstrating the high generalization ability of SCC alongside strong baselines. SCC achieves superior OOD classification performance without losing ID performance. SCC improves the zero-shot performance in terms of AUC from 0.84 to 0.87 on B2 and from 0.73 to 0.77 on B3. After fine-tuning on B2 and B3, the AUC increased from 0.94 to 0.95 on B2 and from 0.78 to 0.83 on B3. These results suggest that SCC can help build robust and generalizable classification models under limited datasets when transferring from models based on natural images.

	\begin{table}[!htbp]
		\caption{\label{tab:performanceBreast} \textbf{Performances of the breast datasets.} AUC of experiments based on the three breast ultrasound datasets and the star (*) stands for our proposed method. Without losing ID performance, SCC has the highest zero-shot and few-shot AUC scores, which implies that it can help build robust and generalizable classification models under limited datasets when transferring from models based on natural images.}
		\centering
		\begin{tabular}{lllll}
			\toprule
			
			Dataset & Method & In-distribution & OOD(0\%) & OOD(100\%)\\
			
			\midrule
			\multirow{4}{*}{B1} & ResNet50 & 0.92  $\pm$ 0.01& \multirow{4}{*}{\textbackslash}	& \multirow{4}{*}{\textbackslash}  \\
			& SimCLR			& 0.92 $\pm$ 0.01 & 									&  \\	
			& DCE* 			& 0.92 $\pm$ 0.01 &  									&  \\ 
			& SCC*	& 0.92 $\pm$ 0.02 &  									&  \\
			\midrule 
			\multirow{4}{*}{B2} & ResNet50 & \multirow{4}{*}{\textbackslash} & 0.84 $\pm$ 0.02 & 0.94  $\pm$ 0.01\\
			& SimCLR			&								  		& 0.84 $\pm$ 0.02 & 0.94  $\pm$ 0.01\\		
			& DCE*				&										& 0.86 $\pm$ 0.03 & 0.94  $\pm$ 0.02\\ 
			& SCC*			&								  		& \textbf{0.87 $\pm$ 0.01} & \textbf{0.95 $\pm$ 0.02} \\
			\midrule 
			\multirow{4}{*}{B3} & ResNet50 & \multirow{4}{*}{\textbackslash} & 0.73  $\pm$ 0.07& 0.78 $\pm$ 0.07  \\
			& SimCLR			&	 								  	& 0.75 $\pm$ 0.04 & 0.80  $\pm$ 0.07\\	
			& DCE* 				&	 								  	& 0.76  $\pm$ 0.02& 0.77 $\pm$ 0.06 \\ 
			& SCC*				&										& \textbf{0.77 $\pm$ 0.02} & \textbf{0.83 $\pm$ 0.02} \\
			\bottomrule 
		\end{tabular}	
	\end{table}

\end{document}